\documentclass[]{interact}

\usepackage{epstopdf}
\usepackage[caption=false]{subfig}
\usepackage{algorithm}
\usepackage{algpseudocode}
\usepackage{amsmath}
\usepackage{graphicx}
\usepackage{lscape}
\usepackage{multirow}
\usepackage{float}
\usepackage{makecell}
\usepackage{array}
\renewcommand\arraystretch{1.2}
\makeatletter
\newenvironment{breakablealgorithm}
  {
   \begin{center}
     \refstepcounter{algorithm}
     \hrule height.8pt depth0pt \kern2pt
     \renewcommand{\caption}[2][\relax]{
       {\raggedright\textbf{\ALG@name~\thealgorithm} ##2\par}%
       \ifx\relax##1\relax 
         \addcontentsline{loa}{algorithm}{\protect\numberline{\thealgorithm}##2}%
       \else 
         \addcontentsline{loa}{algorithm}{\protect\numberline{\thealgorithm}##1}%
       \fi
       \kern2pt\hrule\kern2pt
     }
  }{
     \kern2pt\hrule\relax
   \end{center}
  }
\makeatother

\usepackage{amssymb}
\usepackage[longnamesfirst,sort]{natbib}
\bibpunct[, ]{(}{)}{;}{a}{,}{,}
\renewcommand\bibfont{\fontsize{10}{12}\selectfont}

\theoremstyle{plain}

\theoremstyle{definition}

\theoremstyle{remark}

\begin{document}

\articletype{ARTICLE}

\title{Identifying Fixation and Saccades in Virtual Reality}

\author{
\name{Xiao-lin Chen\textsuperscript{a, c}\thanks{CONTACT Wen-jun Hou. Email: hou1505@163.com} and Wen-jun Hou\textsuperscript{b,c}}
\affil{\textsuperscript{a}School of Automation, Beijing University of Posts and Telecommunications, Beijing, China, cxl95@163.com; \textsuperscript{b}School of Digital Media and Design Arts, Beijing University of Posts and Telecommunications, Beijing, China;
\textsuperscript{c}Beijing Key Laboratory of Network Systems and Network Culture, Beijing, China}
}
\maketitle

\begin{abstract}
Gaze recognition can significantly reduce the amount of eye movement data for a better understanding of cognitive and visual processing. Gaze recognition is an essential precondition for eye-based interaction applications in virtual reality. However, the three-dimensional characteristics of virtual reality environments also pose new challenges to existing recognition algorithms. Based on seven evaluation metrics and the Overall score (the mean of the seven normalized metric values), we obtain optimal parameters of three existing recognition algorithms (Velocity-Threshold Identification, Dispersion-Threshold Identification, and Velocity \& Dispersion-Threshold Identification) and our modified Velocity \& Dispersion-Threshold Identification algorithm. We compare the performance of these four algorithms with optimal parameters. The results show that our modified Velocity \& Dispersion-Threshold Identification performs the best. The impact of interface complexity on classification results is also preliminarily explored. The results show that the algorithms are not sensitive to interface complexity.
\end{abstract}


\section{Introduction}
The essence of eye-movement behavior is the allocation of human attentional resources, no matter whether it is active or passive. One of the main topics of eye movement research is to infer brain activity by monitoring eye movements. Eye-tracking technology provides vital technical support for a deeper understanding of human eye-movement behaviors and the underlying psycho-cognitive activities. Humans mainly have six eye movement types: fixations, saccades, smooth pursuits, optokinetic reflex, vestibulo-ocular reflex, and vergence \citep{2007Leigh}. It is crucial to identify these basic eye movement types from noisy and often inaccurate raw eye position signals for researchers who use eye trackers in their studies. Fixation and saccades are most frequently studied among all six eye movement types, especially in human intention recognition and cognition state recognition\citep{2010Istance}. Fixation identification translates from raw eye-movement data points to fixation locations and implicitly the saccades between them. Fixation identification significantly reduces the raw data size by removing slight eye movements that occur during fixation with little significance in higher-level analysis (such as tremors, drifts, and flicks \citep{1969Alpern, 1980Ditchburn}) and merging raw fixation points into a single representative tuple. Therefore, fixation identification can reduce the noise and volume of raw data while retaining its most essential features to understand cognitive and visual processing behavior.

Virtual reality devices have become more portable and affordable in recent years. Many commercial products have equipped eye-tracking devices. Eye movement, especially fixation, is a natural and intuitive way to interact with the environment. It indicates where our attention is or what we will do next. It is also a part of human nonverbal communication and regulates interaction\citep{2014Pivi}. For example, in collaborative tasks, gaze information can improve communication even more than verbal information by supporting a visual basis\citep{2013Darren}. Therefore, as an input modality of virtual reality, eye tracking can achieve a new and more seamless interaction mode. It enables virtual reality applications to respond to users’ attention and even users’ emotions\citep{2008Brennan}.  Eye-tracking has a long history of application in virtual reality human-computer interaction and has five main applications: user avatar eye behavior simulation\citep{2020Duchowski, 2012Andrist, 2008Queiroz, 2008Lance}, fovea rendering (apply high quality rendering only in the gaze area to reduce power consumption)\citep{2016Weier, 2016Swafford, 2017Roth, 2017Albert}, mitigation of the side effects of vergence-accommodation conflict\citep{2016Gregory, 2014Duchowski, 2013Fisker, 2014Bernhard}, gaze-based interaction (to reduce head movement or improve interaction efficiency)\citep{2019Sidenmark, 2018Rajanna, 2017Piumsomboon, 2017Pfeuffer, 2015Sidorakis}, and user intent or behavior recognition\citep{2021Brendan, 2020Pfeiffer, 2019Alghofaili}. Many of these studies are based on fixation identification.

One of the prerequisites for the wider application of eye-tracking in virtual reality is to recognize the viewpoint and its three-dimensional coordinates based on the sampled eye-tracking data. However, vast existing algorithms on fixation identification are based on the data collected from conventional 2D screens with statistic stimuli, which may not suit Virtual Reality (VR) 3D environments. Because the distribution of fixation points is expanded from two-dimensional to three-dimensional, it becomes more complex to locate the user’s fixation point coordinates. Mobile eye-tracking devices, such as Tobii Glasses for ”the real reality”, are based on video streams taken by a front camera for gaze annotation, which is still essentially analyzing data in a 2D environment, and the accuracy of fixation identification cannot be guaranteed\citep{2012Anneli}.

 \citet{2002Duchowski} first present a velocity-based eye movement analysis algorithm in three-dimensional spaces, applicable to the 3D eye movement data in a virtual environment. They mainly solve the mapping from original 2D eye movement data to 3D world coordinates. However, there is a lack of reasonable evaluation methods. The authors try to compare with some experimental conclusions in the traditional environment. Specifically, the authors find that the average fixation duration is 1.9s in virtual reality. It is significantly different from that in reading (150-650 msec) in previous studies conducted in reality. However, it is difficult to explain whether the difference comes from the virtual reality environment itself or the algorithm’s error. \citet{2013Diaz} present methods identifying fixation and saccadic eye movements in a virtual environment based on the research of \citet{2002Duchowski}, including the calculation of gaze-in-world angles, angular distance from gaze to an object in the virtual world, and algorithms for the identification of pursuit eye movements. Different approaches for fixation identification in 3D scenes have been described by Pelz and Canosa \citet{2001Jeff}, \citet{2006Reimer}, and \citet{2008Munn}. However, these approaches are for monocular eye trackers. Although they can identify fixation in 3D environments, they provide only a fixation direction instead of a 3D fixation position which is important in practical application.

The study of \citet{2020Llanes} develops a dispersion-threshold identification algorithm for data obtained from an eye-tracking system in a head-mounted display. Rule-based criteria are proposed to calibrate the thresholds of the algorithm through different features. However, the difference in the depth of field is not considered in the design of stimuli in their research. Stimuli are presented on two planes with a fixed distance from the user. Secondly, there is no accuracy metric of fixation coordinates to indicate whether the predicted fixation coordinates are consistent with those guided by the stimulus. Furthermore, they also lack a horizontal comparison of different algorithms.

In this paper, based on the three existing gaze classification algorithms, a modified algorithm is proposed to classify fixation and calculate its three-dimensional coordinates. The best parameters of the four classification algorithms, including our algorithm, are obtained through a variety of evaluation metrics. The classification results of each algorithm are from two tasks ( occlusion and non-occlusion) and compared. Overall, our algorithm’s performance is the best. It can identify the user’s actual fixation position, with a velocity threshold of 140°/s, a minimum fixation duration of 130ms, and a dispersion threshold of 5.75° as the optimal parameters. The main contributions of this paper are as follows:

\begin{itemize}
    \item Existing evaluation metrics and classification algorithms are adapted to virtual reality environments to calculate each algorithm’s optimal parameters.
    \item The m-IVDT algorithm is proposed to improve the accuracy of fixation coordinates.
    \item The four algorithms have no preference for interface complexity.
\end{itemize}

The paper is organized as follows. Section 2 reviews the three existing algorithms and introduces the proposed algorithm. Section 3 provides a standardized evaluation system for in-depth quantitative and qualitative analysis of classification results in VR. Section 4 describes our experiment platform and method. Section 5 provides a comparative analysis of the four algorithms. Section 6 concludes our work and suggests future directions.

\section{Algorithm}
\label{sec2}
Fixation-identification algorithms can be based on velocity, dispersion, or area depending on the spatial criteria\citep{2000Salvucci}. Area-based algorithms identify points within given areas of interest (AOIs) representing relevant visual targets. These algorithms provide higher-level assignment of fixation to AOIs, representing higher attentional focus levels on display. Fixation is used as inputs to AOI algorithms. Our research goal is low-level fixation recognition, so area-based algorithms are not in our consideration. Velocity-based algorithms take advantage of the fact that saccades are rapid movements compared to fixation. The most representative velocity-based algorithm is Velocity-Threshold Identification (I-VT), the simplest method to understand and implement. Dispersion-based algorithms emphasize the dispersion of fixed points because they usually are near each other. For example, Dispersion-Threshold Identification (I-DT) identifies fixation as groups of consecutive points within a particular dispersion or maximum separation. We also choose a hybrid algorithm based on these two algorithms, Velocity \& Dispersion-Threshold Identification (IVDT), which integrates speed and discreteness into fixation classification.

This section describes these three algorithms. Sample algorithms are formalized to represent the essential ideas of each class of algorithms and express their basic techniques as simply as possible.

\subsection{Velocity-Threshold Identification}
I-VT begins by calculating point-to-point velocities for each eye data sample. Each velocity is computed by dividing the visual angle between two adjacent points by the time duration.  

$$v_i =\frac{\arccos{\frac{V_i \cdot V_{i+1}}{\parallel V_i \parallel \parallel V_{i+1} \parallel}}}{| t_{i+1} - t_i|} \times 5.73 \times 10^4$$

where $V_i$ is the normalized gaze direction vector at time $t_i$, $V_{i+1}$ is the normalized gaze direction vector at time $t_{i+1}$, $5.73 \times 10^4$ converts the unit from radians per microsecond to degrees per second.  I-VT then classifies each point as a fixation or saccade point based on a simple velocity threshold: if the point’s velocity is lower than the threshold, it is a fixation point; otherwise, it is a saccade point. The process then merges consecutive fixation points into fixation groups. Finally, I-VT translates each fixation group to a tuple $(x_f,y_f,z_f,t_{start},d)$ using the centroid (i.e., the center of mass) coordinates of the points as $x$, $y$, and $z$, the time of the first point as $t$, and the duration of the points as $d$. Algorithm \ref{alg::IVT} presents pseudocode for this I-VT algorithm.

\renewcommand{\algorithmicrequire}{\textbf{Input:}}  
\renewcommand{\algorithmicensure}{\textbf{Output:}} 

\begin{breakablealgorithm}
  \caption{Velocity-Threshold Identification} 
  \label{alg::IVT}
  \begin{footnotesize}
  \begin{algorithmic}[1]
    \Require
      $p_i$: 3D gaze position with timestamps $(x,y,z,t)$;
      $V_i$: normalized gaze direction vector with timestamps $(a,b,c,t)$;
      $Vel$: velocity threshold;
    \Ensure
      $f_i$: representative coordinates corresponding to fixations groups, the starting time and duration of these fixations groups, $(x_f,y_f,z_f,t_{start},d)$
    \State  $//$ calculate instantaneous visual angle
    \For{$i = 0 \to n-1$}
        \State $v_i =\frac{\arccos{\frac{V_i \cdot V_{i+1}}{\parallel V_i \parallel \parallel V_{i+1} \parallel}}}{| t_{i+1} - t_i|} \times 5.73 \times 10^4$
    \EndFor
    \State  \textbf{Initialize} fixation group
    \For{$i = 0 \to n-2$ }
        \If{$v_i < Vel$} \State $V_i$  is added to the fixation group
         \Else 
            \If{ the fixation group not empty} 
                \State Calculate the centroid coordinates $(x_f,y_f,z_f)$ of points in fixation group
                \State Save the timestamp $t$ of the first point in fixation group as $t_{start}$ 
                \State Calculate the duration $d$ of points in fixation group 
                \State  \textbf{Initialize} fixation group
            \EndIf
        \EndIf
    \EndFor
  \end{algorithmic}
\end{footnotesize}
\end{breakablealgorithm}

\subsection{Dispersion-Threshold Identification}
The previous I-DT algorithm uses a sliding window that spans consecutive data points to check for potential fixations. This method is very useful for eye movement data with stable sampling frequency. However, in a virtual reality environment, due to increasing graphic rendering requirements or limited computing power of GPUs, the data collection frequency is unstable and often reduced. Since the raw data is obtained using the SDK (SRanpial) through a Unity script, the data collection frequency depends on the graphic engine’s processing speed. 

To solve this problem, we adjust the algorithm. Instead of setting the initial window, we check whether it meets the minimum fixation duration after determining a group of fixation points. In addition, the dispersion distance between the centroids of two adjacent fixation groups is also checked. If they are too close (below the dispersion threshold), they are merged.

The distance metric we choose is the centroid distance method. The distance is represented by the visual angle between the current point and the following (or previous) point. We only check the distance of the new point to be added to the centroid. If the dispersion is below the dispersion threshold, we expand the fixation group and recalculate the centroid. If the dispersion is above the dispersion threshold, the new point does not correspond to a fixation. Then, we check whether each fixation group meets the minimum fixation duration and whether the dispersion distance from adjacent fixation groups meets the maximum dispersion distance. If both are met, it is regarded as a fixation at the centroid $(x,y,z)$ of the fixation group points with the timestamp of the first point as the fixation start timestamp and the duration of the points as the fixation duration. This process is applied to the entire dataset. Algorithm \ref{alg::IDT} presents pseudocode for this I-DT algorithm.

\begin{breakablealgorithm}
  \caption{Dispersion-Threshold Identification} 
  \label{alg::IDT}
  \begin{footnotesize}
  \begin{algorithmic}[2]
    \Require
      $p_i$: 3D gaze position with timestamps $(x,y,z,t)$;
      $DD_{max}$: maximum fixation dispersion distance threshold;
      $Duration_{min}$: minimum fixation duration threshold;
    \Ensure
      $f_i$: representative coordinates corresponding to fixations groups, the starting time and duration of these fixations groups, $(x_f,y_f,z_f,t_{start},d)$
    \State  \textbf{Initialize} Previous fixation group $PFG$ and Current fixation group $CFG$
    \State save $p_0$ into $PFG$
    \State save $p_1$ into $CFG$
    \For{$i = 2 \to n-1$}
        \State Calculate the $CFG$ centroid coordinates $(x,y,z)$ 
        \State Calculate the dispersion distance ($DD$) between $CFG$ centroid coordinates and $p_i$ coordinates
        \If{$DD<DD_{max}$} \State save $p_i$ into $CFG$
        \Else
            \If{ $CFG$ is not empty} \State Calculate the duration $d$ of points in $CFG$
                \If{$d>Duration_{min}$}  \State Calculate the dispersion distance ($DD$) between first point in $CFG$ and last point in $PFG$
                    \If{$DD<DD_{max}$} \State Marge $CFG$ into $PFG$
                    \Else 
                        \State Calculate the $PFG$ centroid coordinates $(x_f,y_f,z_f)$ 
                        \State Save the timestamp $t$ of the first point in $PFG$ as $t_{start},$ 
                        \State Calculate the duration $d$ of points in $PFG$ 
                        \State  \textbf{Initialize} $PFG$
                        \State Marge $CFG$ into $PFG$
                        \State  \textbf{Initialize} $CFG$
                        \State save $p_i$ into $CFG$
                    \EndIf
                \Else
                    \State  \textbf{Initialize} $CFG$
                    \State save $p_i$ into $CFG$
                \EndIf
            \EndIf
        \EndIf
    \EndFor
  \end{algorithmic}
 \end{footnotesize}
\end{breakablealgorithm}

\subsection{Velocity \& Dispersion-Threshold Identification}
\citet{2013Komogortsev} propose a ternary classification algorithm called velocity and dispersion threshold identification (IVDT). It first identifies saccades by the velocity threshold. Subsequently, it identifies smooth pursuits from fixation by a modified dispersion threshold and duration.

The original algorithm still needs an initial time window to carry out, and smooth pursuit is not one of our classification categories, so we modify the algorithm.

\begin{figure}[H]
\centering
\includegraphics[width=12cm]{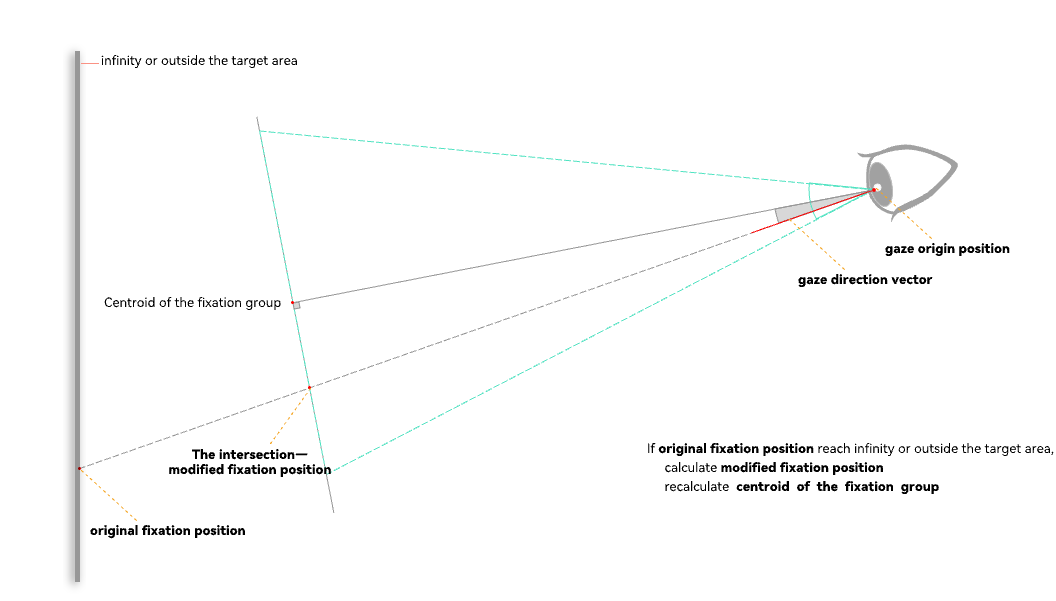} 
\caption{The intersection is the modified sampling point.} \label{fig:1}
\end{figure}

The I-VDT algorithm in this paper still employs three thresholds of velocity, dispersion, and minimum fixation duration. Same as I-VT, I-VDT begins by calculating point-to-point velocities for each eye data sample. Then I-VDT classifies(Algorithm \ref{alg::IVDT}) each point as a fixation or saccade point based on a simple velocity threshold: if the point’s velocity is below the threshold, it is a fixation point; otherwise, it is a saccade point. Then, we check whether each fixation group meets the minimum fixation duration and whether the dispersion distance from adjacent fixation groups meets the maximum dispersion distance. If both are met, it is regarded as a fixation at the centroid  $(x,y,z)$ of the fixation group points with the timestamp of the first point as fixation start timestamp and the duration of the points as fixation duration.

We use the gaze-based ray-casting method to calculate gaze-object intersections as gaze intersection points to show where the participant is looking at. However, when the line of sight deviates for a short time, the Z coordinates of gaze intersection points may be very different. That is to say, the user’s actual gaze does not change much. However, in the 3D gaze-based ray-casting method, the Z coordinates of gaze points are quite different. In this case, the centroid method in two-dimensional interfaces may cause a large error on the z-axis, so we propose a modified method (m-IVDT) to calculate the centroid of the fixation group.

The basic idea of this method is to transform the coordinates of the sampling points whose Z coordinates reach infinity or outside the target area (in this experimental environment, if Z is greater than or equal to 4.9, the sampling point collides with the wall which is the farthest from the user in the virtual room). As shown in Figure \ref{fig:1}, firstly, we take the direction between the pupil position and the existing fixation as the normal vector and construct the plane through the fixation. Then we calculate the intersection of the plane and the line of sight formed by the pupil position and the sampling point. The intersection is the modified original sampling point to recalculate the centroid of the fixation group, i.e., the new fixation coordinate.

\begin{breakablealgorithm}
  \caption{Velocity \& Dispersion-Threshold Identification} 
  \label{alg::IVDT}
  \begin{footnotesize}
  \begin{algorithmic}[3]
    \Require
        $p_i$:3D gaze position with timestamps, $(x,y,z,t)$; 
        $V_i$:normalized gaze direction vector with timestamps,
        $Vel$:velocity threshold;
        $DD_{max}$: maximum fixation dispersion distance threshold;
        $Duration_{min}$: minimum fixation duration threshold;
    \Ensure
    $f_i$:representative coordinates corresponding to fixations groups, the starting time and duration of these fixations groups, $(x_f,y_f,z_f,t_{start},d)$
      \State  $//$ calculate the instantaneous visual angle
    \For{$i = 0 \to n-1$}
        \State $v_i =\frac{\arccos{\frac{V_i \cdot V_{i+1}}{\parallel V_i \parallel \parallel V_{i+1} \parallel}}}{| t_{i+1} - t_i|} \times 5.73 \times 10^4$
    \EndFor
    \State  \textbf{Initialize} Previous fixation group $PFG$ and Current fixation group $CFG$
    \State save $p_0$ into $PFG$
    \State save $p_1$ into $CFG$
    \For{$i = 2 \to n-1$}
        \State Calculate the $CFG$ centroid coordinates $(x,y,z)$ 
        \State Calculate the dispersion distance ($DD$) between $CFG$ centroid coordinates and $p_i$ coordinates
        \If{$v_i < Vel$}  \State save $p_i$ into $CFG$
        \Else
            \If{ $CFG$ is not empty} \State Calculate the duration $d$ of points in $CFG$
                \If{$d>Duration_{min}$}  \State Calculate the dispersion distance ($DD$) between the first point in $CFG$ and the last point in $PFG$
                    \If{$DD<DD_{max}$} \State Marge $CFG$ into $PFG$
                    \Else 
                        \State Calculate the $PFG$ centroid coordinates $(x_f,y_f,z_f)$ 
                        \State Save the timestamp $t$ of the first point in $PFG$ as $t_{start}$ 
                        \State Calculate the duration $d$ of points in $PFG$ 
                        \State  \textbf{Initialize} $PFG$
                        \State Marge $CFG$ into $PFG$
                        \State  \textbf{Initialize} $CFG$
                        \State save $p_i$ into $CFG$
                    \EndIf
                \Else
                    \State  \textbf{Initialize} $CFG$
                    \State save $p_i$ into $CFG$
                \EndIf
            \EndIf
        \EndIf
    \EndFor
  \end{algorithmic}
  \end{footnotesize}
\end{breakablealgorithm}

\section{Evaluation}

\citet{2010Komogortsev} define a set of qualitative and quantitative scores to assess classification algorithms’ performance. For fixation and saccade classification algorithm, they propose seven evaluation metrics: the average number of fixation (ANF), average fixation duration (AFD), the average number of saccades (ANS), and average saccade amplitude (ASA) as four well-known metrics, and fixation quantitative score (FQnS), fixation qualitative score (FQlS), and saccade quantitative score (SQnS) as three original metrics. The scores originally measure the classification quality when only fixation and saccades are present in a two-dimensional environment’s raw eye positional trace. We perform the following slight modifications to extend behavior scores for a three-dimensional virtual reality environment.

\subsection{Fixation quantitative score (FQnS)}
\begin{figure}[H]
\centering
\includegraphics[width=12cm]{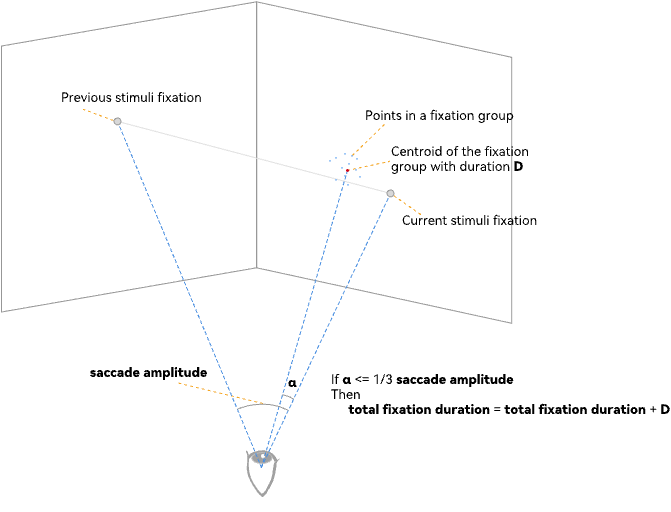} 
\caption{Total fixation duration.} \label{fig:2}
\end{figure}

FQnS compares the amount of detected fixation behavior to the actual amount of fixation behavior encoded in the stimuli\citep{2010Komogortsev}. Suppose the original recorded eye-positional signal is classified as fixation with its centroid in spatial proximity of the stimulus fixation, which is 1/3 of the amplitude of the previous stimulus saccade(Figure \ref{fig:2}), the total fixation duration is incremented by the duration of the fixation group.

FQnS is calculated by normalizing the total resulting fixation duration by the actual total duration of fixation points encoded in the stimulus.

$$ FQnS =100\% \times \frac{total\;fixation\;duration}{stimuli\;total\;fixation\;duration}$$

Ideal FQnS never reaches 100\% because it takes time for the central nervous system to send a neuronal signal to relevant muscles to execute a saccade\citep{2007Leigh}. The beginning of fixation is always delayed by 200ms plus the duration of a saccade\citep{2007Leigh}. Therefore, ideal FQnS is calculated by the following equation:
$$D_{sacDur_j} = (2.2 \times A_{sacAmp_j}+21)$$

$$Ideal\_FQnS = 100\% \times (1-\frac{m\times S_l + \sum_{j=1}^m D_{sacDur_j}}{\sum_{i=1}^n D_{stimFixDur_i}})$$

where $m$ is the number of stimulus saccades; $S_l$ is the saccadic latency of 200ms; $A_{sacAmp_j}$ is the saccade’s amplitude of the $j_th$ stimulus saccade measured in degrees; $D_{sacDur_j}$ is the expected duration of the $j_th$ stimulus saccade; $n$ is the number of stimulus fixation; $D_{stimFixDur_i}$ is the duration of the $i_th$ stimulus fixation.

\subsection{Fixation qualitative score (FQlS)}
FQlS compares the spatial proximity of the classified eye-fixation signal to the actual stimulus signal, indicating the positional accuracy or error of the classified fixation. FQlS is calculated with the same formula proposed by \citet{2010Komogortsev}. If a sampled eye point is classified as fixation, it calculates the Euclidean distance between the fixation group’s centroid coordinates $(x_c,y_c,z_c)$ and the corresponding stimulus fixation coordinates $(x_s,y_s,z_s)$. Then the average of these distances is calculated as follows:

$$fixationDistance_i=\sqrt{(x_c-x_s)^2+(y_c-y_s)^2+(z_c-z_s)^2} $$
$$FQlS = \frac{1}{N} \sum_{i=1}^N fixationDistance_i$$

where $N$ is the number of sampled points classified as fixation.

In \citet{2010Komogortsev}, it is assumed that the ideal value of FQlS is about $0.5$ degrees because the accuracy of modern eye trackers is generally less than $<0.5$ degrees. It is assumed that the distance between human eyes and the two-dimensional interface is constant, and the transformation between visual angle and Euclidean distance is easy to compute. However, in a three-dimensional environment, Euclidean distance cannot be directly transformed to the visual angle. In the actual test, the accuracy of virtual reality eye trackers is still lower than that of traditional eye trackers. According to the preliminary analysis of the prediction data, we hypothesize that the practical value of FQlS should be around $0.5$, and the unit is the same as the Euclidean distance.                                                                                       
\subsection{Saccade quantitative score (SQnS)}
SQnS is calculated with the same formula proposed by \citet{2010Komogortsev}.Two separate numbers need to be computed. The first one represents the amount of stimulus saccadic behavior, i.e., ”total stimuli saccade amplitude”. The second represents the amount of classified saccadic behavior, i.e., ”total detected saccade amplitude”. SQnS is computed by the following equation:

$$SQnS =100\% \times \frac{total\;detected\;saccade\;amplitude}{total\; stimuli\;saccade\;amplitude}$$

An SQnS of 100\% indicates that the integral sum of detected eye saccade amplitudes equals that of the actual stimuli. So a closer SQnS to 100\% denotes better performance.

\section{Experiment}
\subsection{Participant}
A total of 11 participants (six females and five males), ages 22-27 years with an average age of 24 (±1.53), were recruited. All participants have a normal or corrected-to-normal vision. Eligible for participation in the experiment was only healthy people who did not have any cognitive or motor impairments. None of the participants reported known visual or vestibular disorders, such as color or night blindness, a displacement of balance. Ten had corrected vision who used glasses or lenses during the experiment. Nine had tried HMDs several times before, and two had no prior VR experience. Three had prior experience with eye-based interaction.
\subsection{Apparatus}
Participants were instructed to wear an HTC Vive Pro Eye with one headset and a built-in eye tracker. The headset had a resolution of 1440 ×1600 pixels/eye, and 2880 ×1600 pixels were combined with a 110°field of view. The headset's highest refresh rate was 90 Hz. The refresh rate of the built-in eye tracker was 120Hz, which offers a tracking precision of 0.5°-1.1°. The experience was conducted on a PC with an Intel Core i7-9700, an NVIDIA GeForce GTX 1070 8G GPU, and 16G DDR4 2666Hz RAM. The experimental platform was developed using Unity 2019.4 and C\#. 
\subsection{Procedure}
\label{sec4.3}
The experiment takes approximately 10 minutes in total for each participant. Each participant is given a brief introduction of the purposes and objectives of the experiment before signing a consent form. Participants are asked to sit in a natural sitting position and keep head position fixed as much as possible during the experiment, but heads’ natural rotation is allowed.

The virtual stimuli consist of a virtual room with the participant in the center. At the beginning of the experiment, the eye tracker is recalibrated for accuracy by asking the participant to gaze at targets at five varying points on display. The calibration process takes approximately 1 minute.

There are two kinds of stimuli. One stimulus presents a blue sphere in the center of participants’ vision in the virtual room. It then stays in that position for 1.5s before changing position. The position of the sphere is changed 20 times in each session. Each position is generated randomly in a cube space with a center point $(0,1.2,2.2)$ and a side length of 1.6. Only one sphere is displayed on the scene at any time. Participants are asked to gaze at the sphere during the whole session. The other stimulus has 19 blue spheres randomly displayed in the same cube space as the first one. Besides, another sphere in the center of the participant’s vision is the start sphere. The spheres turn red in random order from the central one. Each red sphere remains red for 1.5 seconds and then returns to blue. It is repeated until all 20 spheres turn red once. There is always one red sphere and 19 blue spheres displayed on the scene at any time. Participants are asked to gaze at the red sphere during the whole session.

Either of these two stimuli is repeated five times, so an experiment consists of ten sessions, and the order of the ten sessions is random. Participants have no break time between sessions. The target sphere display area is limited to ensure that the participants can see the target without moving their heads widely.

\subsection{Data set}
\begin{figure}[H]
\centering
\includegraphics[width=10cm]{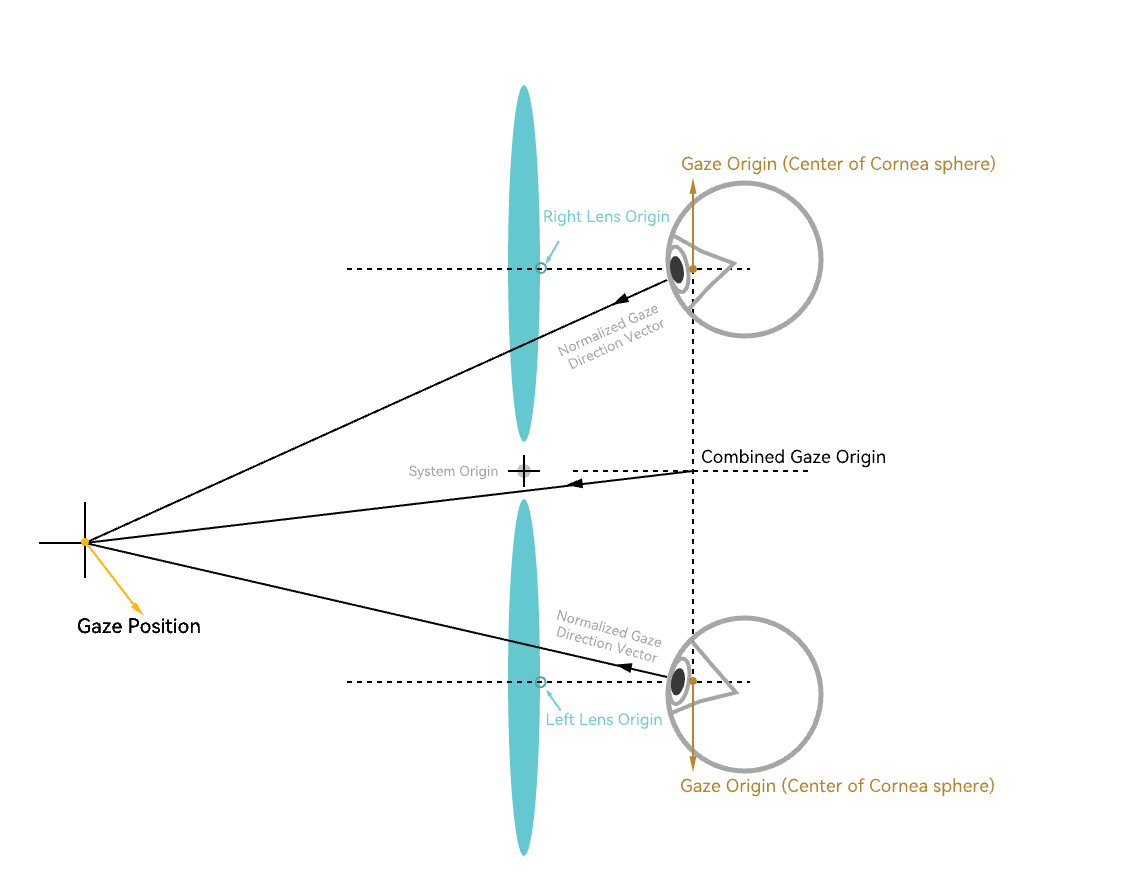} 
\caption{Eye tracker data output} \label{fig:Dataset}
\end{figure}

Our eye tracker data include users’ combined gaze origin positions, combined normalized gaze direction vectors, pupil diameters, eye openness, head-set positions, and corresponding timestamps (Figure \ref{fig:Dataset}). Eleven subjects conduct a total of 110 sessions. After removing invalid data, 100 valid sessions are obtained with 168075 raw data samples. 

Data preprocessing includes interpolating the missing data and coordinate system transformation. The main reason for missing data is blinking. The majority of missing data are Gaze Original and Gaze Direction Normalized. We choose to fill the missing data with the last valid data. The reason is that in our follow-up research, the whole classification pipeline, including data preprocessing, should be able to run in real-time. There are 4205 missing data samples, accounting for about 2.5\% of the data set. The raw data is obtained using the SDK (SRanpial) through a Unity script. According to the documentation of SRanpial, Gaze Original is the point in the eye from which the gaze ray originates, and Gaze Direction Normalized is the normalized gaze direction of the eye. They are both based on a right-handed coordinate system. However, Unity is based on a left-handed coordinate system. Therefore, we multiply their X-axis coordinates by -1 to convert the right-handed coordinate system to a left-handed one. Secondly, Gaze Original is based on the eye position, i.e., the main camera’s position in the three-dimensional environment, so a further conversion is needed, which adds the coordinates of the main camera to Gaze Original.

In a virtual reality environment, the geometry of the presented stimuli is known. 3D gaze positions can be inferred by calculating the intersection point of the gaze direction vector and the reconstructed 3D scene with a ray-based approach \citep{2001Duchowski,2002Duchowski,2016Mansouryar}. A gaze direction vector and the corresponding gaze original position are used to find the point of intersection with the reconstructed 3D scene representing the 3D gaze point.

\subsection{Threshold Tuning}
It is important to test the performance of each classification algorithm over a sensible range of threshold values.

In two influential reviews of research on eye movements in information processing, \citet{1992Rayner,1998Rayner} reports that the mean fixation durations vary with tasks, such as 225–250 ms for (silent) reading, 180–275 ms for visual search, and 260–330 ms for scene viewing. As for saccade amplitude, the mean fixation durations also vary with tasks, such as about 2° for reading, 3° for visual search, and 4° for scene viewing. The research of \citet{1999Andrews} also gives a similar conclusion. They report that the average fixation durations for reading are 150–220 ms, for visual search are 190–230 ms, and for scene viewing are 200–400 ms. Saccade sizes vary from 3° to 7° during scene viewing, 3° to 8° during reading, and 4° to 7° during visual search. A velocity threshold of 130°/s and a minimum fixation duration of 150ms are suggested by \citet{2002Duchowski}.The minimum fixation duration should be less than the average fixation duration of each task in these studies, and the maximum saccade amplitude should be less than the average saccade amplitude of each task. 

Based on these previous studies, for I-DT and I-VDT, the dispersion threshold range is set from 1.0 degrees sto 6.0degrees with a step size of 0.25degrees, and the minimal fixation duration threshold is set from 50ms to 150ms with a step size of 10ms. The range of velocity threshold values for I-VT and I-VDT is set from 30degrees/s to 150degrees/s with a step size of 10degrees/s.

We use the grid search method to traverse all parameter combinations. The three algorithms classify each session’s data, and the seven evaluation metrics are calculated respectively. Considering the simple stimulus behavior and the normal subject pool, the following metrics are set up as ideal metric performance: $Ideal\_AFN = 21$ fixations, $Ideal\_AFD = 1.5 s$, $Ideal\_ASA = 20$ saccades, $ Ideal\_FQlS = 0.5$, and $Ideal\_SQnS = 100\%$. Because the positions of 20 stimuli spheres in each session are random, the ideal ASA in each session is different and needs to be calculated separately. The calculation of the ideal value of FQnS is also related to the angle between stimuli spheres and needs to be calculated separately. Theoretically, the closer to the ideal value, the better the algorithm’s performance in this metric, so we use the absolute difference between the actual value and the ideal value to express the algorithm’s performance. The unit of each metric is not the same. To make a better comparative analysis, min-max normalization transforms the absolute difference of each metric to $[0,1]$.  

$$y_{normalized} = \frac{y-y_{min}}{y_{max}-y_{min}}$$

The Overall score is the average value of the normalized scores of each metric and is taken as the final comprehensive performance score. The optimal parameters are selected according to this score.

\section{Result and Discussion}
\subsection{Tuning Parameter Values for Fixation-identification Algorithms}
Parameter values for these algorithms greatly influence their output, so a direct comparison between these algorithms has to be done with caution. Because of the change of interactive environment (virtual reality), most of the parameter values of these algorithms cannot be applied from the literature directly. Hence, it is necessary to tune the parameter values for the particular environment. 

\subsubsection{Tuning Parameter Values for Velocity-Threshold Identification}

A one-way ANOVA examines the impact of velocity threshold on all seven metrics and the Overall score for Velocity-Threshold Identification. Table \ref{tab:1} shows the ANOVA analysis with statistical significance. The significance value is below 0.05 for the Overall score, FQnS, FN, SN, and SQnS. Therefore, there is statistical significance in these five metrics between different velocity thresholds chosen. However, there is no significant difference in FQlS, FAD, or ASA.

\begin{table}[H]
\setlength{\abovecaptionskip}{0cm}
\setlength{\belowcaptionskip}{0.2cm}
\tiny
\centering
\caption{One-Way ANOVA Output of Velocity-Threshold of IVT on all seven metrics and the Overall score}
\label{tab:1}
\resizebox{\textwidth}{!}{%
\renewcommand{\arraystretch }{1.5}
\begin{tabular}{llrrrrr}
\hline
\multicolumn{1}{c}{} & \multicolumn{1}{c}{} & \multicolumn{1}{c}{Sum of Squares} & \multicolumn{1}{c}{df} & \multicolumn{1}{c}{Mean Square} & \multicolumn{1}{c}{F} & \multicolumn{1}{c}{Sig.} \\ \hline
FQnS                 & Between Groups       & 6.441                              & 12                     & 0.537                           & 52.641                & 0.000                    \\
                     & Within Groups        & 12.989                             & 1274                   & 0.01                            &                       &                          \\
                     & Total                & 19.43                              & 1286                   &                                 &                       &                          \\
FQlS                 & Between Groups       & 0.04                               & 12                     & 0.003                           & 0.14                  & 1.000                    \\
                     & Within Groups        & 30.56                              & 1274                   & 0.024                           &                       &                          \\
                     & Total                & 30.6                               & 1286                   &                                 &                       &                          \\
FN                   & Between Groups       & 43.059                             & 12                     & 3.588                           & 259.412               & 0.000                    \\
                     & Within Groups        & 17.622                             & 1274                   & 0.014                           &                       &                          \\
                     & Total                & 60.681                             & 1286                   &                                 &                       &                          \\
AFD                  & Between Groups       & 4.239                              & 12                     & 0.353                           & 183.296               & 0.000                    \\
                     & Within Groups        & 2.456                              & 1274                   & 0.002                           &                       &                          \\
                     & Total                & 6.695                              & 1286                   &                                 &                       &                          \\
SN                   & Between Groups       & 42.621                             & 12                     & 3.552                           & 259.408               & 0.000                    \\
                     & Within Groups        & 17.443                             & 1274                   & 0.014                           &                       &                          \\
                     & Total                & 60.064                             & 1286                   &                                 &                       &                          \\
ASA                  & Between Groups       & 0.001                              & 12                     & 0                               & 0.003                 & 1.000                    \\
                     & Within Groups        & 41.62                              & 1274                   & 0.033                           &                       &                          \\
                     & Total                & 41.621                             & 1286                   &                                 &                       &                          \\
SQnS                 & Between Groups       & 4.291                              & 12                     & 0.358                           & 33.651                & 0.000                    \\
                     & Within Groups        & 13.537                             & 1274                   & 0.011                           &                       &                          \\
                     & Total                & 17.828                             & 1286                   &                                 &                       &                          \\
Overall Score        & Between Groups       & 7.745                              & 12                     & 0.645                           & 85.513                & 0.000                    \\
                     & Within Groups        & 9.609                              & 1273                   & 0.008                           &                       &                          \\
                     & Total                & 17.354                             & 1285                   &                                 &                       &                          \\ \hline
\end{tabular}%
}
\end{table}

\begin{figure}[H]
\centering
\includegraphics[width=12cm]{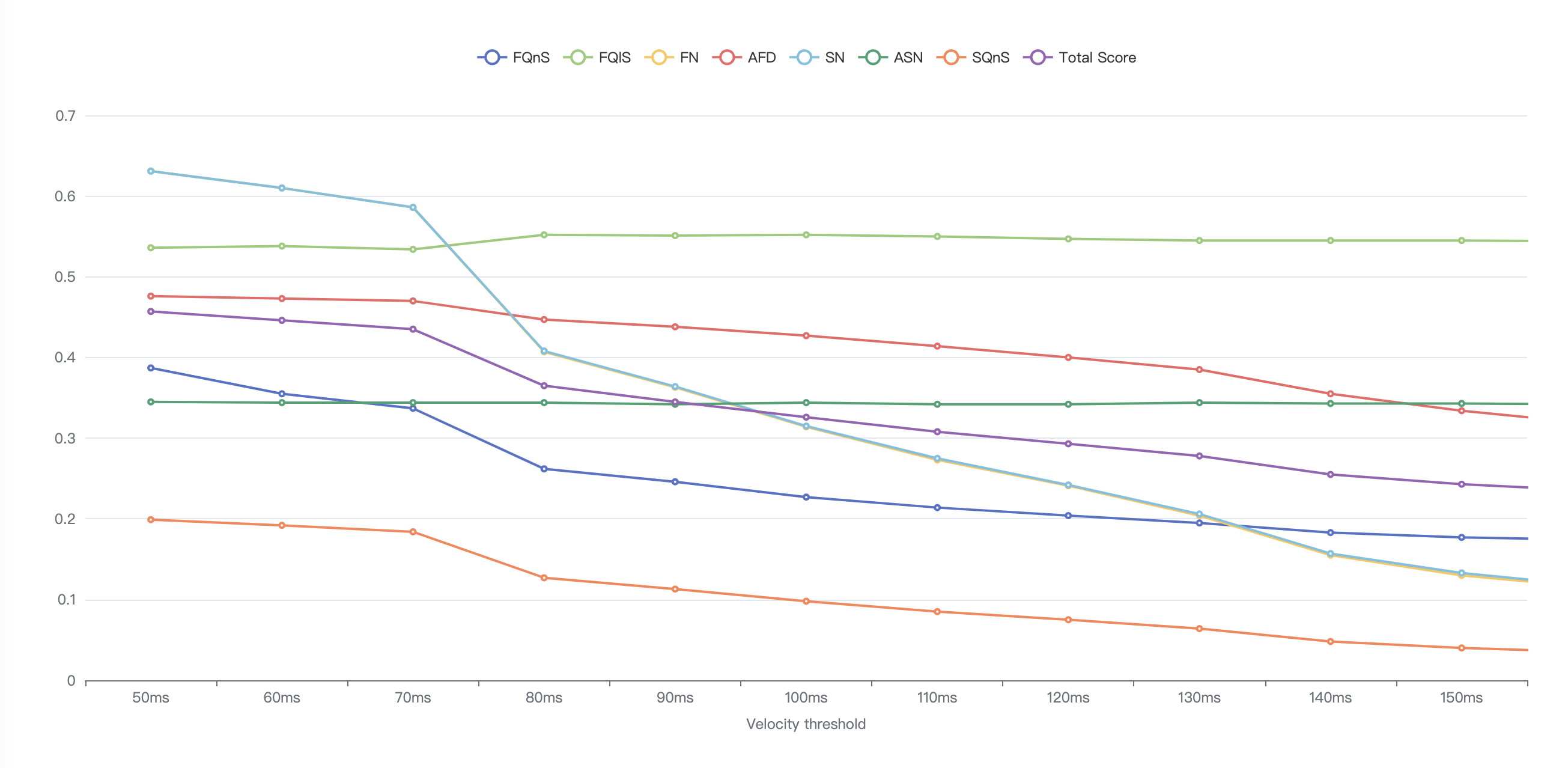} 
\caption{Line chart of velocity thresholds and all eight metrics} 
\label{fig:IVT}
\end{figure}

It can be more clearly seen from Figure \ref{fig:IVT} that with the increase of velocity threshold, the five evaluation metric values decrease with significant differences, i.e., the classification result is closer to the ideal result, so 150 is the optimal parameter for the velocity threshold.

\subsubsection{Tuning Parameter Values for Dispersion-Threshold Identification}
A two-way ANOVA examines the impact of minimum fixation duration and maximum dispersion angle on the Overall score. There is no statistically significant interaction between minimum fixation duration and maximum desperation angle impacting the Overall score of classification, F (190, 21772) = 0.626, p = 1.000. Further note that partial eta squared is only 0.005 for our interaction effect, which is negligible. Two-way ANOVAs examine the impact of minimum fixation duration and maximum dispersion angle on all seven metrics. There is a statistically significant interaction between minimum fixation duration and maximum dispersion angle impacting FQnS, FN, SN, ASA, and SQnS, but not FQlS or AFD. We use line charts to more intuitively present the impact of parameters on various metrics

\begin{figure}[H]
\centering
\includegraphics[width=12cm]{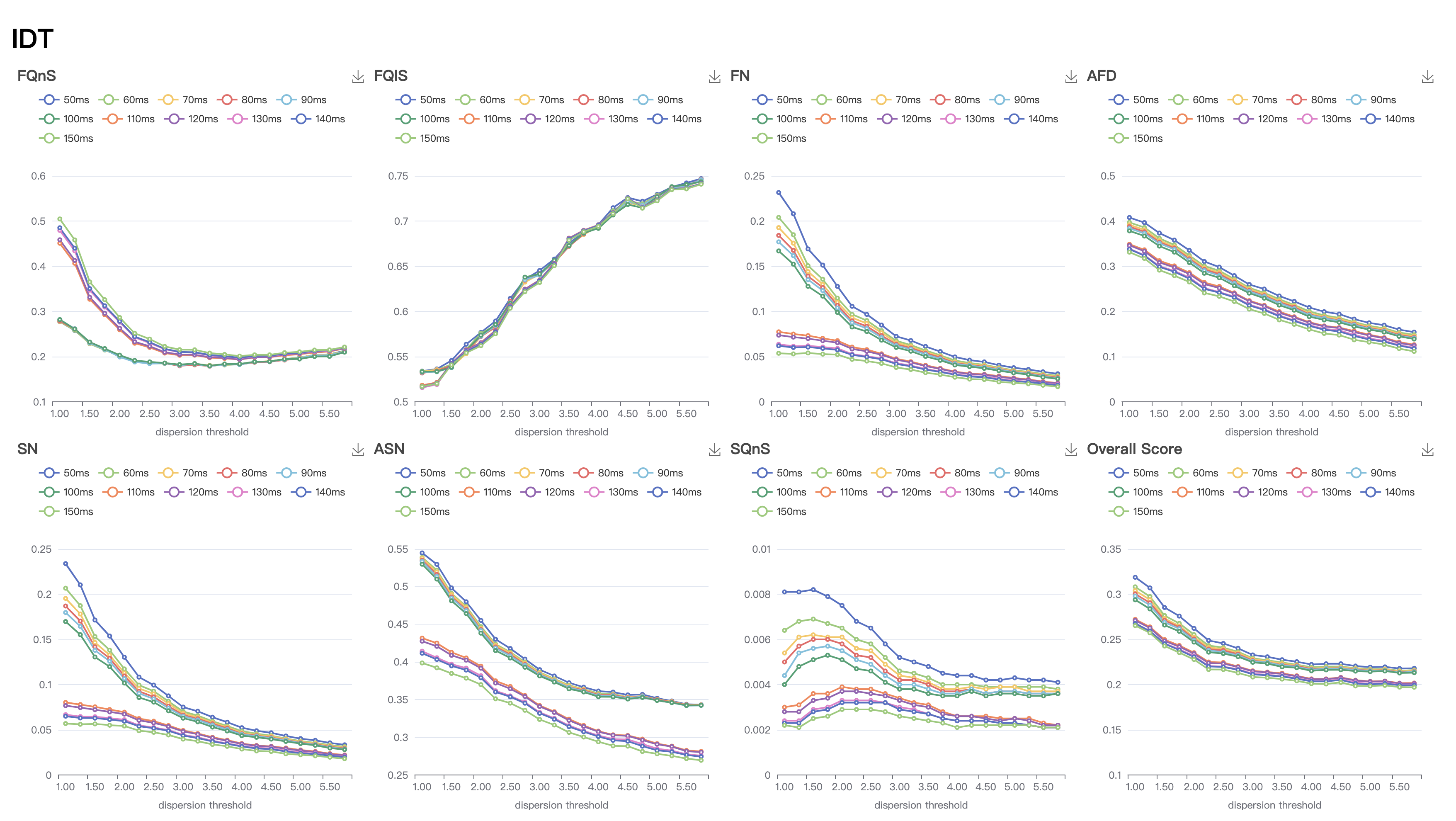} 
\caption{Line charts of IDT parameters and all eight metrics} 
\label{fig:IDT}
\end{figure}

As can be seen from the line charts (Figure \ref{fig:IDT}), except FQlS, all metric values decrease with the increase of dispersion threshold and max dispersion angle. As for the minimum fixation duration, FN, AFD, Sn, ASA, SQnS, and Overall score decrease with the increase of minimum fixation duration. Based on the results of each metric, we choose the minimum fixation duration of 150ms and the dispersion threshold of 5.75° as the optimal parameters of IDT.

\subsubsection{Tuning Parameter Values for Velocity \& Dispersion-Threshold Identification}
IVDT (Velocity \& Dispersion-Threshold Identification) algorithm includes two algorithms: IVDT and m-IVDT algorithms. Based on the classification results of these two algorithms, three-way ANOVAs examine the impact of velocity threshold, minimum fixation duration, and maximum dispersion angle on the Overall score and other seven metrics. This paper mainly takes the Overall score for detailed analysis.

\begin{table}[]
\setlength{\abovecaptionskip}{0cm}
\setlength{\belowcaptionskip}{0.2cm}
\centering
\caption{Three-Way ANOVA Output of three parameter of IVDT on all seven metrics and the Overall score}
\label{tab:2}
\resizebox{\textwidth}{!}{%
\renewcommand{\arraystretch }{1.5}
\begin{tabular}{llllll}
\hline
Source                                           & \multicolumn{1}{c}{Type III Sum of Squares} & \multicolumn{1}{c}{df} & \multicolumn{1}{c}{Mean Square} & \multicolumn{1}{c}{F} & \multicolumn{1}{c}{Sig.} \\ \hline
Corrected Model                                  & 714.340a                                    & 2859                   & 0.25                            & 39.648                & 0.000                    \\
Intercept                                        & 18994.174                                   & 1                      & 18994.174                       & 3014022.546           & 0.000                    \\
Vel\_threshold                                    & 87.296                                      & 12                     & 7.275                           & 1154.358              & 0.000                    \\
min\_fix\_dur                                      & 2.015                                       & 10                     & 0.202                           & 31.977                & 0.000                    \\
max\_angle                                        & 555.574                                     & 19                     & 29.241                          & 4639.963              & 0.000                    \\
Vel\_threshold * min\_fix\_dur                      & 7.575                                       & 120                    & 0.063                           & 10.016                & 0.000                    \\
Vel\_threshold * max\_angle                        & 40.726                                      & 228                    & 0.179                           & 28.344                & 0.000                    \\
min\_fix\_dur * max\_angle                          & 17.658                                      & 190                    & 0.093                           & 14.747                & 0.000                    \\
Vel\_threshold * min\_fix\_dur * max\_angle          & 3.563                                       & 2280                   & 0.002                           & 0.248                 & 1.000                    \\
Error                                            & 1781.532                                    & 282696                 & 0.006                           &                       &                          \\
Total                                            & 21498.379                                   & 285556                 &                                 &                       &                          \\
Corrected Total                                  & 2495.872                                    & 285555                 &                                 &                       &                          \\ \hline
a.   R Squared = .286 (Adjusted R Squared = .279) &                                             &                        &                                 &                       &                         
\end{tabular}%
}
\end{table}

For the classification results of IVDT, a three-way ANOVA examines the impact of velocity threshold, minimum fixation duration, and maximum dispersion angle on the Overall score. There is no statistically significant interaction between velocity threshold, min fixation duration, and max dispersion angle impacting the Overall score of classification, F (2280, 282696) = 0.248, p = 1.000. There is a statistically significant interaction between velocity threshold and minimum fixation duration, velocity threshold and maximum dispersion angle, and maximum fixation duration and maximum dispersion angle impacting the Overall score. The simple main effects are analyzed for each group of factors with interactive influence. Except when the maximum angle is in [3,4], the impact of different minimum fixation durations on the Overall score is not statistically significant. Other analysis results show that the impact of various factors on the Overall score is statistically significant. Through the line chart (Appendix \ref{sec:sample:appendixA}), we can more intuitively show the impact of various factors on different indicators. Based on the analysis results of each metric, we choose the velocity threshold of 140 °/s, the minimum fixation duration of 110ms, and the dispersion threshold of 5.75° as the optimal parameters of IVDT.

\begin{table}[]
\setlength{\abovecaptionskip}{0cm}
\setlength{\belowcaptionskip}{0.2cm}
\centering
\caption{Three-Way ANOVA Output of three parameter of m-IVDT on all seven metrics and the Overall score}
\label{tab:3}
\resizebox{\textwidth}{!}{%
\renewcommand{\arraystretch }{1.5}
\begin{tabular}{llllll}
\hline
Source                                           & \multicolumn{1}{c}{Type III Sum of Squares} & \multicolumn{1}{c}{df} & \multicolumn{1}{c}{Mean Square} & \multicolumn{1}{c}{F} & \multicolumn{1}{c}{Sig.} \\ \hline
Corrected Model                                  & 905.943a                                    & 2859                   & 0.317                           & 57.385                & 0.000                    \\
Intercept                                        & 11709.328                                   & 1                      & 11709.328                       & 2120517.371           & 0.000                    \\
Vel\_threshold                                    & 93.144                                      & 12                     & 7.762                           & 1405.664              & 0.000                    \\
min\_fix\_dur                                      & 3.809                                       & 10                     & 0.381                           & 68.984                & 0.000                    \\
max\_angle                                        & 726.464                                     & 19                     & 38.235                          & 6924.211              & 0.000                    \\
Vel\_threshold * min\_fix\_dur                      & 6.745                                       & 120                    & 0.056                           & 10.179                & 0.000                    \\
Vel\_threshold * max\_angle                        & 51.42                                       & 228                    & 0.226                           & 40.842                & 0.000                    \\
min\_fix\_dur * max\_angle                          & 20.197                                      & 190                    & 0.106                           & 19.25                 & 0.000                    \\
Vel\_threshold * min\_fix\_dur * max\_angle          & 4.226                                       & 2280                   & 0.002                           & 0.336                 & 1.000                    \\
Error                                            & 1561.025                                    & 282696                 & 0.006                           &                       &                          \\
Total                                            & 14183.474                                   & 285556                 &                                 &                       &                          \\
Corrected Total                                  & 2466.968                                    & 285555                 &                                 &                       &                          \\ \hline
a.   R Squared = .367 (Adjusted R Squared = .361) &                                             &                        &                                 &                       &                         
\end{tabular}%
}
\end{table}

For the classification results of m-IVDT, a three-way ANOVA examines the impact of three parameters on the Overall score. There is no statistically significant interaction between velocity threshold, minimum fixation duration, and maximum dispersion angle impacting the Overall score of classification, F (2280, 282696) = 0.336, p = 1.000. There is a statistically significant interaction between velocity threshold and minimum fixation duration, velocity threshold and maximum dispersion angle, and maximum fixation duration and maximum dispersion angle impacting the Overall score. The simple main effects are analyzed for each group of factors with interactive influence. The results are the same as the analysis of IVDT, except when the maximum angle is in [3,4], the impact of different minimum fixation durations on the Overall score is not statistically significant. Other analysis results show that the impact of various factors on the Overall score is statistically significant. Through the line chart chart(Appendix \ref{sec:sample:appendixB}), we can more intuitively show the impact of various factors on different indicators. Based on the analysis results of each metric, we choose the velocity threshold of 140°/s, the minimum fixation duration of 130ms, and the dispersion threshold of 5.75° as the optimal parameters of m-IVDT.

\subsection{Comparison of four algorithms}

We treat every session as an independent test and calculate seven metrics and the Overall score of the classification results of the four algorithms (with optimal parameters) for each session. One-way ANOVA determines whether different algorithms impact the seven metrics and the Overall score.

Table \ref{tab:4}  shows the output of the ANOVA analysis with statistical significance. The significance value is below 0.05 for all eight metrics. Therefore, there is a statistically significant difference in all eight metrics between the different algorithms.

\begin{table}[]
\setlength{\abovecaptionskip}{0cm}
\setlength{\belowcaptionskip}{0.2cm}
\centering
\caption{One-Way ANOVA Output of four algorithms on all seven metrics and the Overall score}
\label{tab:4}
\resizebox{\textwidth}{!}{%
\renewcommand{\arraystretch }{1.5}
\begin{tabular}{lllllllll}
\hline
         & FQnS        & FQlS        & FN          & AFD         & SN          & ASA         & SQnS        & Overall Score \\ \hline
IVT      & 0.172±0.09  & 0.544±0.155 & 0.104±0.063 & 0.305±0.071 & 0.106±0.063 & 0.343±0.175 & 0.032±0.047 & 0.229±0.071 \\
IDT      & 0.221±0.091 & 0.741±0.105 & 0.017±0.016 & 0.111±0.068 & 0.018±0.016 & 0.27±0.08   & 0.002±0.002 & 0.197±0.039 \\
IVDT     & 0.236±0.121 & 0.54±0.155  & 0.008±0.009 & 0.06±0.053  & 0.008±0.01  & 0.246±0.093 & 0.003±0.002 & 0.157±0.042 \\
m-IVDT   & 0.236±0.127 & 0.109±0.104 & 0.008±0.009 & 0.057±0.053 & 0.007±0.009 & 0.246±0.092 & 0.003±0.002 & 0.095±0.034 \\
F(3,393) & 7.623       & 404.85      & 195.809     & 357.2       & 208.213     & 15.469      & 38.72       & 140.213     \\
Sig.     & 0.000       & 0.000       & 0.000       & 0.000       & 0.000       & 0.000       & 0.000       & 0.000       \\ \hline
\end{tabular}%
}
\end{table}

\begin{figure}
\centering
\includegraphics[width=12cm]{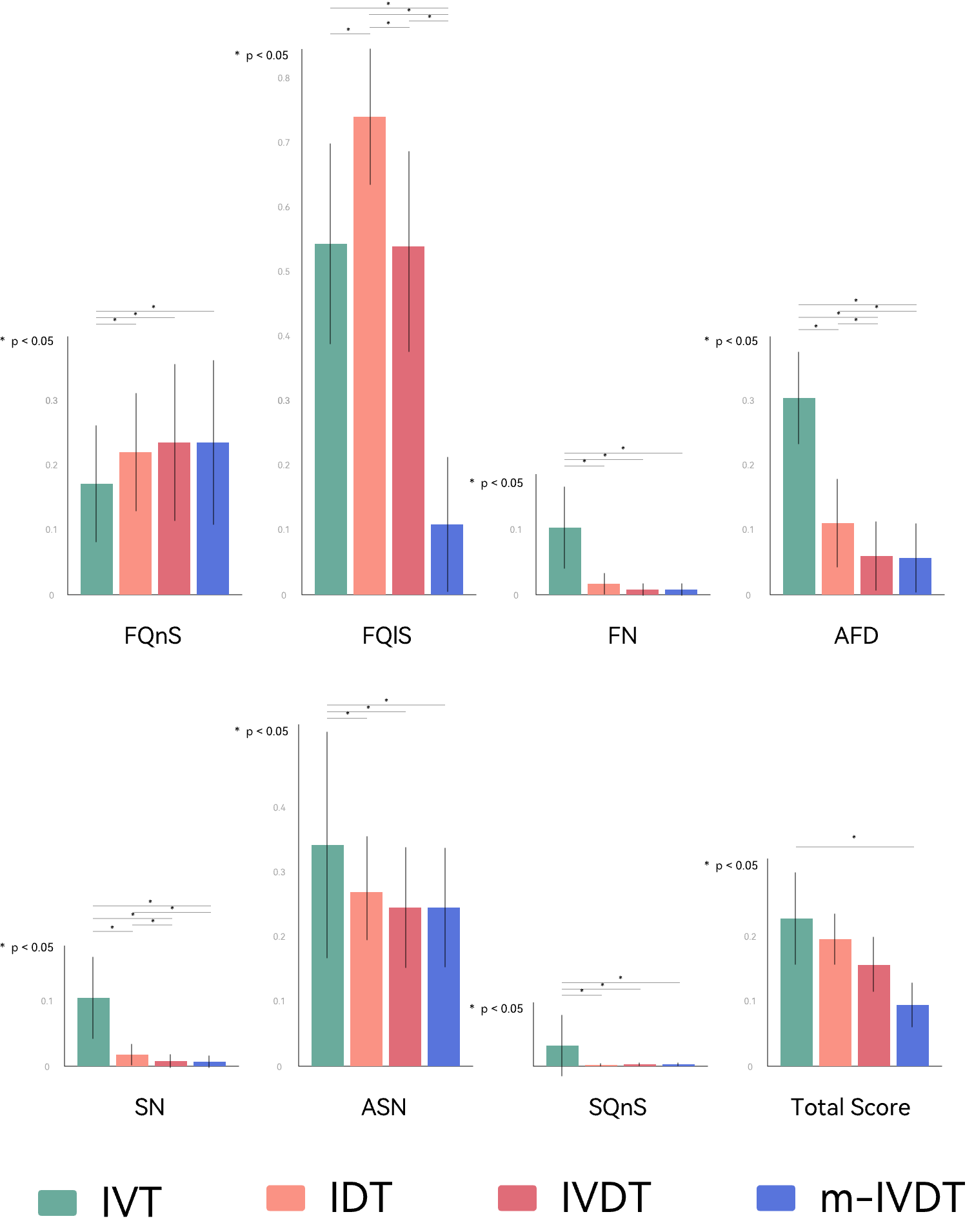} 
\caption{Statistics of the Four Algorithms} 
\label{fig:algorithm}
\end{figure}

An LSD post hoc test reveals that FQnS is statistically significantly higher for IDT (0.221 ± 0.090, p = 0.002), IVDT (0.235 ± 0.121, p = 0.000), and m-IVDT (0.236 ± 0.127, p = 0.000) compared to IVT (0.172 ± 0.090). FQlS is statistically significantly higher for IVT (0.544 ± 0.155, p = 0.000), IDT (0.741 ± 0.105, p = 0.000), and IVDT (0.540 ± 0.155, p = 0.000) compared to m-IVDT (0.109 ± 0.104). FN is statistically significantly higher for IVT (0.104 ± 0.063,p = 0.000) and IDT (0.017 ± 0.016, p = 0.000) compared to IVDT (0.008 ± 0.009) and m-IVDT (0.008 ± 0.009). AFD is statistically significantly higher for IVT (0.305 ± 0.071, p = 0.000) and IDT (0.111 ± 0.068, p = 0.000) compared to IVDT (0.060 ± 0.053) and m-IVDT (0.057 ± 0.053). SN is statistically significantly higher for IVT (0.106 ± 0.063, p = 0.000) and IDT (0.018 ± 0.016, p= 0.000) compared to IVDT (0.008 ± 0.010) and m-IVDT (0.007 ± 0.009). ASA is statistically significantly higher for IVT (0.343 ± 0.175, p = 0.000) and IDT (0.270 ± 0.080, p = 0.000) compared to IVDT (0.246 ± 0.093) and m-IVDT (0.246 ± 0.092). SQnS is statistically significantly lower for IDT (0.002 ± 0.002, p = 0.002), IVDT (0.003 ± 0.002, p = 0.000), and m-IVDT (0.003 ± 0.002, p =0.000) compared to IVT (0.032 ± 0.047). As for Overall score, it is statistically significantly higher for IVT (0.229 ± 0.071, p = 0.000), IDT (0.197 ± 0.039, p = 0.000), and IVDT (0.157 ± 0.042, p = 0.000) compared to m-IVDT (0.095 ± 0.034). Figure \ref{fig:algorithm} shows the statistics of the evaluation results of each algorithm. In conclusion, IVT performs the best in FQnS, followed by IDT. There is no difference between IVDT and m-IVDT. m-IVDT performs the best in FQlS and the worst in IDT. There is no difference between FQlS and IVT, but both perform better than IDT. In FN, AFD, Sn, and ASA, there is no difference between FQlS and m-FQlS. IVT performs the worst in these indicators, followed by IDT. IVT performs the worst in SQnS, and the other three algorithms have no significant difference. As for the Overall score, m-FQlS is the best, followed by FQlS and IDT, and IVT is the worst.

\subsection{Comparison of the two tasks}

We evaluate the algorithms on the two tasks in Section \ref{sec4.3}. The major difference between Tasks 1 and 2 is that Task 2 has multiple objects in the interface simultaneously and only one is the real target, while Task 1 has only one target in the interface at any time. The main purpose is to study whether the interface’s complexity affects the algorithms’ classification results.
\begin{table}[]
\setlength{\abovecaptionskip}{0cm}
\setlength{\belowcaptionskip}{0.2cm}
\small
\centering
\caption{One-Way ANOVA Output of Task type on all seven metrics and the Overall score}
\label{tab:5}
\resizebox{\textwidth}{!}{%
\renewcommand{\arraystretch }{1.5}
\begin{tabular}{lllllllll}
\hline
 & FQnS & FQlS & FN & AFD & SN & ASN & SQnS & Overall Score \\ \hline
Task1 & 0.167±0.099 & 0.5±0.266 & 0.033±0.051 & 0.127±0.115 & 0.034±0.052 & 0.256±0.117 & 0.01±0.027 & 0.161±0.070 \\
Task2 & 0.263±0.101 & 0.469±0.265 & 0.035±0.053 & 0.139±0.122 & 0.035±0.054 & 0.295±0.125 & 0.01±0.025 & 0.178±0.070 \\
F(1,395) & 90.157 & 1.337 & 0.078 & 1.041 & 0.062 & 10.155 & 0.023 & 5.973 \\
Sig. & 0.000 & 0.248 & 0.781 & 0.308 & 0.804 & 0.002 & 0.878 & 0.015 \\ \hline
\end{tabular}%
}
\end{table}

\begin{figure}
\centering
\includegraphics[width=12cm]{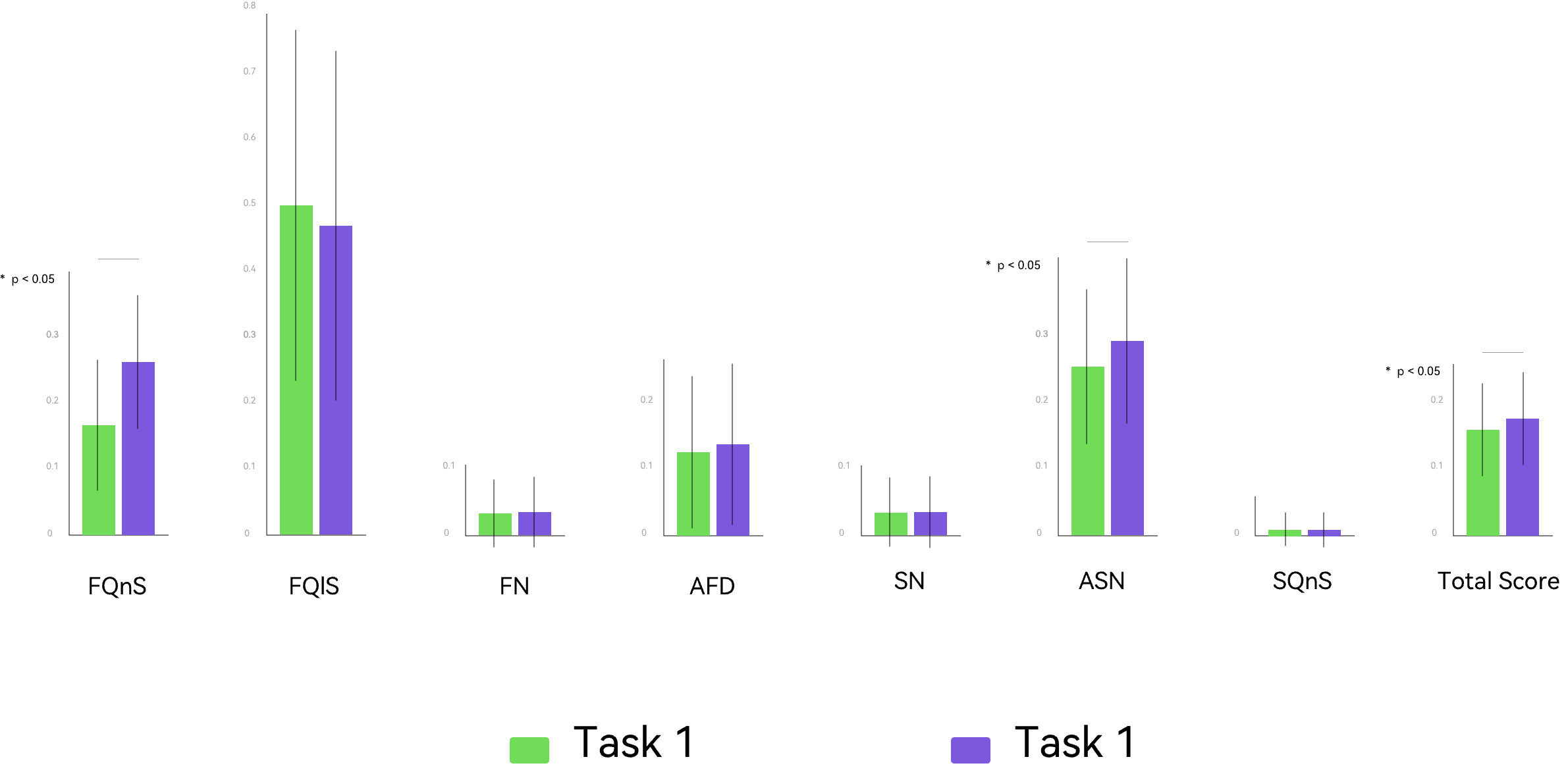} 
\caption{Descriptive Statistic of two tasks} 
\label{fig:task}
\end{figure}

One-way ANOVA determines whether the type of tasks impacts all seven metrics and the Overall score. The result in Table \ref{tab:5} indicats that Task 1 has statistically significantly lower FQnS (0.167 ± 0.167, F(1,395) = 90.157, p = 0.000) and ASA (0.256 ± 0.117, F(1,395) = 10.155, p = 0.002) compared to Task 2 (0.263 ±0.101, 0.295 ± 0.125). Furthermore, the Overall score of Task 2 is 0.178±0.070, statistically significantly different from Task 1 (0.161±0.070, F(1,395) = 5.973, p= 0.015). Figure \ref{fig:task} shows the statistics of the evaluation results of both tasks.

\begin{table}[]
\setlength{\abovecaptionskip}{0cm}
\setlength{\belowcaptionskip}{0.2cm}
\centering
\caption{Two-Way ANOVA Output of Task type and classifier on all seven metrics and the Overall score}
\label{tab:6}
\resizebox{\textwidth}{!}{%
\renewcommand{\arraystretch }{1.5}
\begin{tabular}{clllllllll}
\hline
\multicolumn{2}{c}{} & \textbf{FQnS} & \textbf{FQlS} & \textbf{FN} & \textbf{AFD} & \textbf{SN} & \textbf{ASN} & \textbf{SQnS} & \textbf{Overall Score} \\ \hline
\multirow{3}{*}{\textbf{Task}} & F(1,389) & 96.213 & 5.062 & 0.167 & 3.773 & 0.136 & 11.193 & 0.024 & 8.348 \\
 & Sig. & 0.000 & 0.025 & 0.683 & 0.053 & 0.712 & 0.001 & 0.876 & 0.004 \\
 & Partial Eta Squared & 0.198 & 0.013 & 0.000 & 0.010 & 0.000 & 0.028 & 0.000 & 0.021 \\ \hline
\multirow{3}{*}{\textbf{classifier}} & F(3,389) & 9.403 & 412.224 & 193.823 & 357.124 & 206.046 & 15.774 & 38.279 & 141.516 \\
 & Sig. & 0.000 & 0.000 & 0.000 & 0.000 & 0.000 & 0.000 & 0.000 & 0.000 \\
 & Partial Eta Squared & 0.068 & 0.761 & 0.599 & 0.734 & 0.614 & 0.108 & 0.228 & 0.522 \\ \hline
\multirow{3}{*}{\textbf{Task * classifier}} & F(3,389) & 1.075 & 1.865 & 0.047 & 0.333 & 0.035 & 0.032 & 0.016 & 0.068 \\
 & Sig. & 0.359 & 0.135 & 0.986 & 0.802 & 0.991 & 0.992 & 0.997 & 0.977 \\
 & Partial Eta Squared & 0.008 & 0.014 & 0.000 & 0.003 & 0.000 & 0.000 & 0.000 & 0.001 \\ \hline
\multirow{4}{*}{\textbf{Task1}} & IVT & 0.126±0.071 & 0.574±0.128 & 0.103±0.061 & 0.297±0.067 & 0.106±0.061 & 0.326±0.163 & 0.032±0.049 & 0.223±0.066 \\
 & IDT & 0.186±0.08 & 0.748±0.108 & 0.015±0.013 & 0.102±0.064 & 0.016±0.013 & 0.248±0.07 & 0.002±0.002 & 0.188±0.032 \\
 & IVDT & 0.177±0.109 & 0.571±0.127 & 0.008±0.007 & 0.057±0.049 & 0.008±0.007 & 0.225±0.092 & 0.003±0.002 & 0.150±0.036 \\
 & m-IVDT & 0.18±0.12 & 0.102±0.095 & 0.007±0.007 & 0.054±0.048 & 0.007±0.007 & 0.225±0.091 & 0.003±0.002 & 0.083±0.031 \\ \hline
\multirow{4}{*}{\textbf{Task2}} & IVT & 0.216±0.084 & 0.515±0.173 & 0.104±0.065 & 0.314±0.073 & 0.106±0.065 & 0.359±0.185 & 0.032±0.044 & 0.235±0.076 \\
 & IDT & 0.255±0.088 & 0.734±0.103 & 0.018±0.019 & 0.121±0.071 & 0.019±0.019 & 0.29±0.084 & 0.002±0.002 & 0.206±0.043 \\
 & IVDT & 0.291±0.104 & 0.511±0.173 & 0.008±0.011 & 0.062±0.058 & 0.008±0.011 & 0.266±0.09 & 0.003±0.002 & 0.164±0.045 \\
 & m-IVDT & 0.288±0.111 & 0.116±0.112 & 0.008±0.01 & 0.06±0.057 & 0.008±0.011 & 0.265±0.09 & 0.004±0.002 & 0.107±0.033 \\ \hline
\multicolumn{2}{c}{\textbf{Adjusted   R Squared}} & 0.234 & 0.758 & 0.592 & 0.73 & 0.607 & 0.115 & 0.214 & 0.519 \\ \hline
\end{tabular}%
}
\end{table}

\begin{figure}
\centering
\includegraphics[width=12cm]{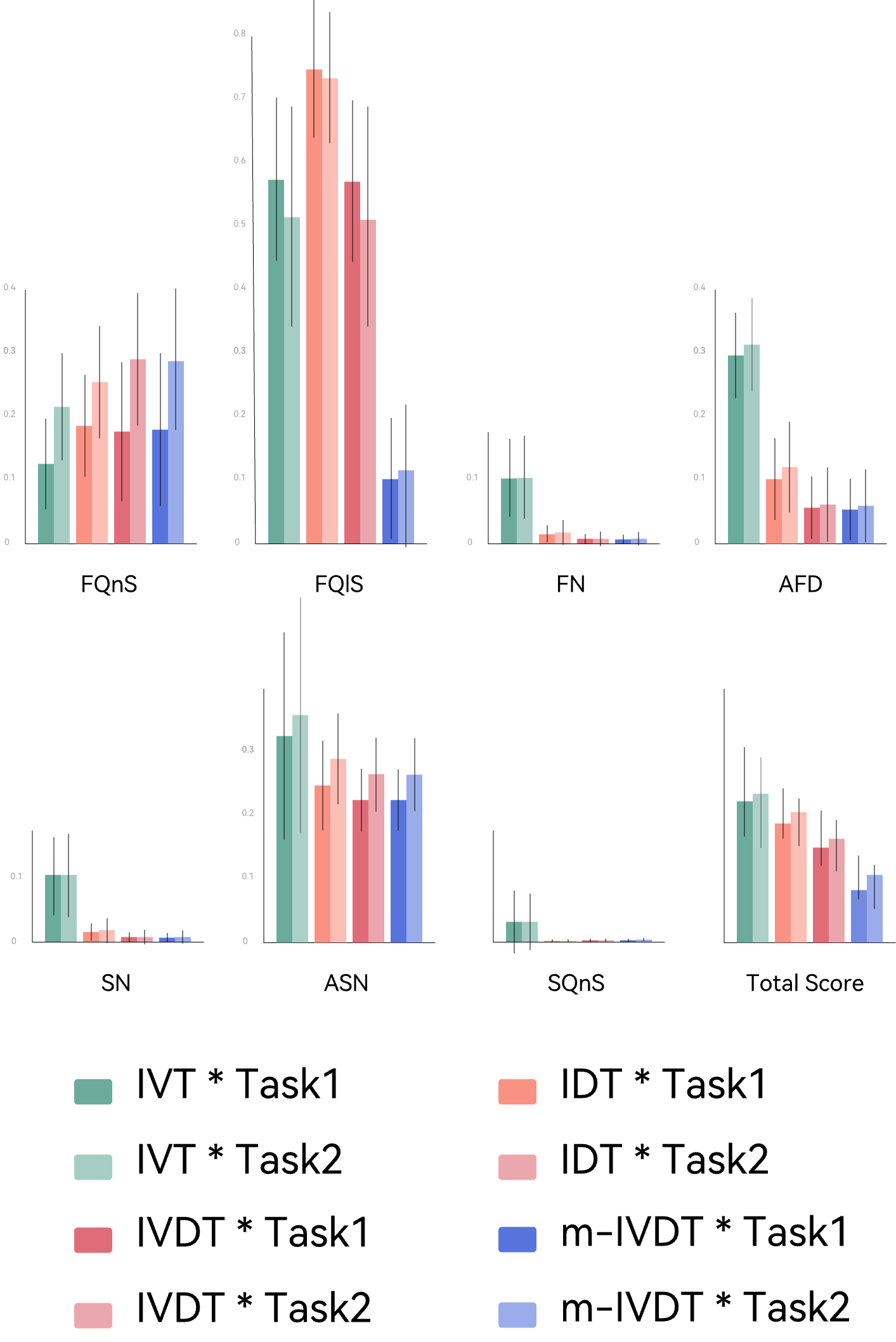} 
\caption{Statistics of the Combinations of the Two Tasks and Four Algorithms} 
\label{fig:taskalg}
\end{figure}

Two-way ANOVA examines the impact of different tasks and classification algorithms on all seven metrics and the Overall score. If both task and algorithm types are considered, four classification algorithms and two tasks generate eight combinations. The interaction between different tasks and classification algorithms for all seven metrics and the Overall score cannot be demonstrated (Table \ref{tab:6}).Therefore, we look into the main effects. As shown in the table, there are significant differences in all eight metrics between algorithms. As for tasks, there are significant differences in FQnS, FQlS, ASN, and Overall score between tasks. Further note that partial eta squares of every metric are lower than 0.015 for the interaction effect, which is negligible. Last but not least, adjusted r squared tells us that the percent of the variance in each matric is attributable to task and algorithm types. Figure \ref{fig:taskalg} shows the statistics of the evaluation results of each combination of task and algorithm.

In general, there is no interaction effect between task types and algorithms, as there is no significant difference in the results of different task types under different algorithms. For the impact of task types on classification results, the main difference is that there are significant differences in FQnS, ASA, and Overall score. The performance on Task 1 is better than that on Task 2 in terms of these metrics, which is consistent with our expectations. Two reasons contribute to it. One is that the complex interface affects users’ actual gaze behavior and makes it deviate from the ideal eye movement trajectory of the stimulus design. The other is that the interface complexity may also affect the classification accuracy of algorithms, but the current experimental design cannot determine which one contributes more. However, the main purpose of this analysis is to explore whether each algorithm is sensitive to interface complexity. The results show that the task type does not affect the classification performance, i.e., the algorithms are not sensitive to interface complexity.

\section{Conclusions and Future Work}

This paper aims to explore the eye movement behavior classification algorithms used in virtual reality environments. The classification algorithm's classification effect was evaluated by comparing the classification results with the ideal eye movement behavior preset by stimuli: FQnS, FQlS, FN, AFD, SN, ASA, sqns, and the average overall score of seven indicators. Firstly, the parameters with the best all-around performance under each algorithm are selected by analyzing the overall score under different parameter combinations. Then compare and analyze the performance of IVT, IDT, IVDT, and m-IVDT algorithms (optimal parameters) in eight indicators. We found that IVT performed best on FQnS; that is, when judging whether the current moment belongs to the fixation point through angular velocity, the accuracy is the highest. However, due to the lack of screening for fixation points with a short duration, there will be too many fixation points, so it performed worst on FN, AFD, Sn, and ASA, related to the number of fixation points. The disadvantage of IDT is that when judging the fixation point only through the spatial position, it will be greatly affected by the error of the original data of spatial coordinates. Therefore, IDT performs the worst in FQlS; the fixation coordinates obtained in the IDT algorithm are the least consistent with the stimulus. The IVDT algorithm combines the advantages of the two to a certain extent; that is, it does not rely too much on space coordinates that are not necessarily completely accurate. At the same time, it can avoid too many fixation points, but it is still difficult to give accurate fixation point coordinates. The m-IVDT proposed by us solves this problem better. By correcting the possible wrong spatial coordinate values, the accuracy of the eye movement fixation coordinate is improved. The classification results of eye movement behavior have certain usability.

The main limitation of this study is that we compare the algorithms’ classification results with stimuli. The implicit assumption is that the actual eye movement behavior of users is consistent with the changes of the stimuli, but it may not be true. We choose the simplest visual environment and the most basic selection task to minimize the impact of other factors on users. In the following research, we can also consider using artificial eye movement data as a standard method to verify our research results, and we can also avoid the influence of stimuli on the classification results. In future research, we will also consider using the deep learning method to classify eye movement behavior and take the differences between individuals into account to get an adaptive algorithm according to the user’s situation.

\section*{Funding}

Thanks to all participants. This work was supported by XXX (Grant No.: XXX).

\bibliographystyle{elsarticle-num-names} 
\bibliography{main}

\appendix
\section{Line chart of IVDT parameters}
\label{sec:sample:appendixA}
\begin{figure}[H]
\centering
\includegraphics[width=0.85\textwidth]{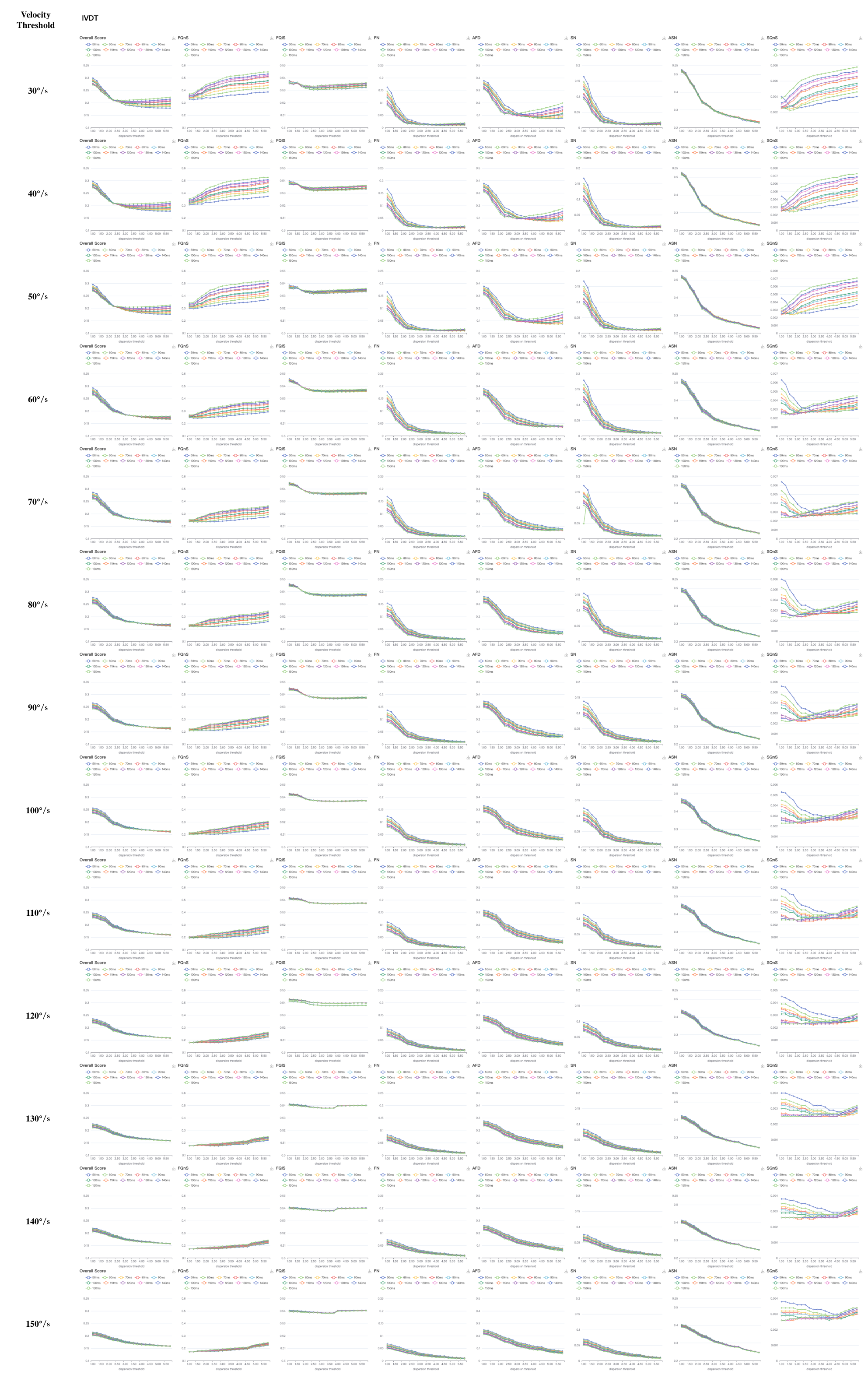} 
\end{figure}

\section{Line chart of m-IVDT parameters}
\label{sec:sample:appendixB}
\begin{figure}[H]
\centering
\includegraphics[width=0.85\textwidth]{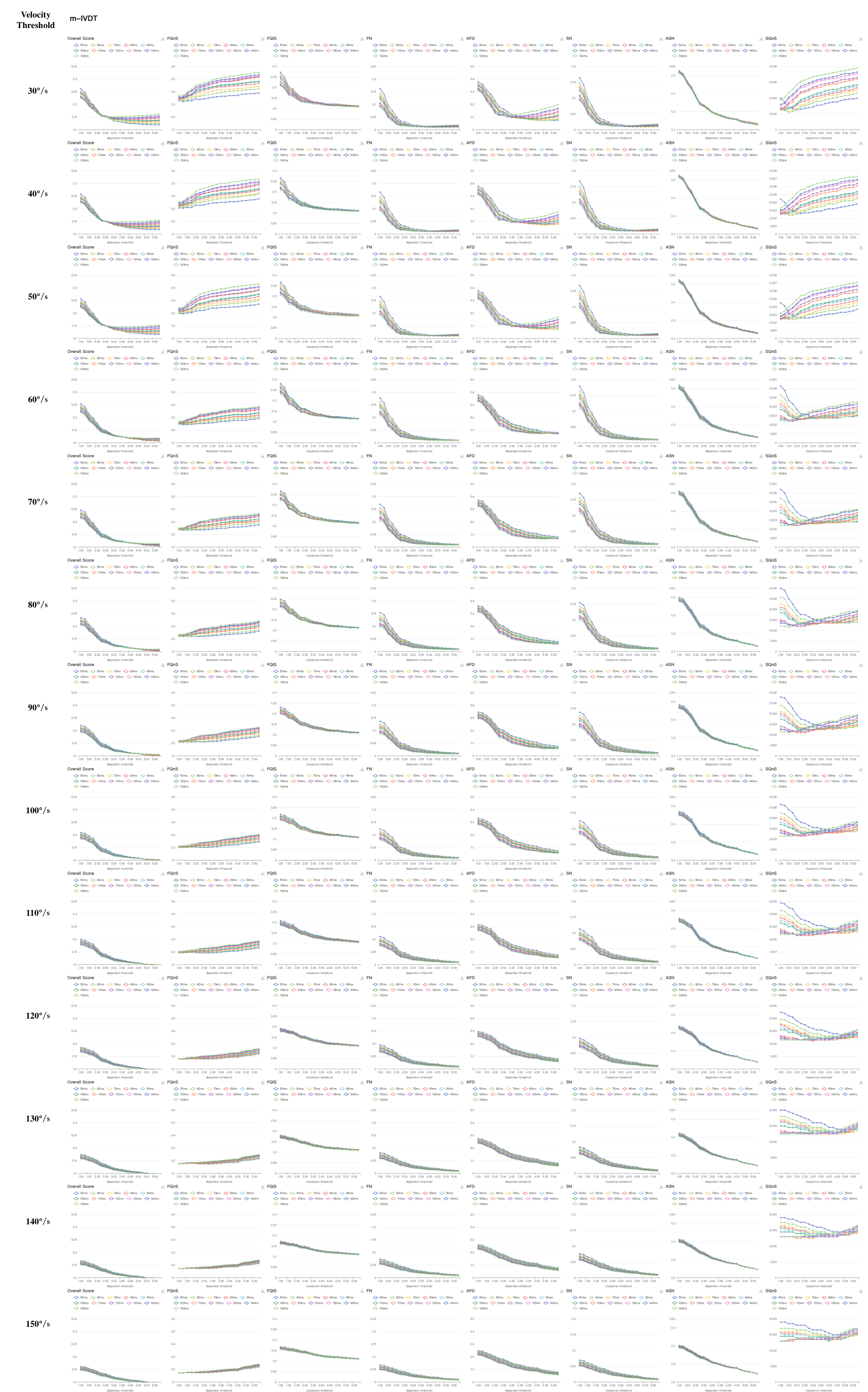} 
\end{figure}

\section{Author Bio}
Wen-jun Hou (Professor, Doctoral supervisor, Vice Chairman UXACN) received BE in ME, Taiyuan University of Science and Technology, Taiyuan, China in June 1991, the Ph.D. degrees from the Beijing University of Posts and Telecommunications, Beijing, China in June 2006. Her research interests include natural interaction, information visualization, interaction experience and VR/AR. Currently, she is the assistant dean at the School of Digital Media \& Design, Beijing University of Post and Telecommunication, Beijing, China.

Xiao-lin Chen received BE in Industrial Design, Beijing University of Posts and Telecommunications, Beijing, China in June 2016. Her research interests include eye-based interaction, voice interaction and user experience. Currently, she is pursuing a Ph.D. degree in Mechatronic Engineering from Beijing University of Post and Telecommunication, Beijing, China.

\end{document}